\documentclass[sigconf]{acmart}
\AtBeginDocument{%
  }

\copyrightyear{2025}
\acmYear{2025}
\setcopyright{rightsretained}
\acmConference[SSTD '25]{19th International Symposium on Spatial and Temporal Data}{August 25--27, 2025}{Osaka, Japan}
\acmBooktitle{19th International Symposium on Spatial and Temporal Data (SSTD '25), August 25--27, 2025, Osaka, Japan}
\acmPrice{}
\acmDOI{10.1145/3748777.3748810}
\acmISBN{979-8-4007-2094-9/25/08}

\begin{document}

\title{LLMs Meet Cross-Modal Time Series Analytics: Overview and Directions}

\author{Chenxi Liu}
\affiliation{%
  \institution{Nanyang Technological University,}
  \country{Singapore}
}
\email{chenxi.liu@ntu.edu.sg}

\author{Hao Miao*}
\affiliation{%
  \institution{The Hong Kong Polytechnic University,}
  \city{Hong Kong}
  \country{China}}
\email{hao.miao@polyu.edu.hk}


\author{Cheng Long}
\affiliation{%
  \institution{Nanyang Technological University,}
  \country{Singapore}
  }
\email{c.long@ntu.edu.sg}

\author{Yan Zhao}
\affiliation{%
  \institution{University of Electronic Science and Technology of China,}
  \country{China}
  }
\email{zhaoyan@uestc.edu.cn}

\author{Ziyue Li}
\affiliation{%
  \institution{Technical University of Munich,}
  \country{Germany}}
\email{zlibn@connect.ust.hk}

\author{Panos Kalnis}
\affiliation{%
  \institution{King Abdullah University of Science and Technology,}
  \country{Saudi Arabia}
  }
\email{panos.kalnis@kaust.edu.sa}

\thanks{* Hao Miao is the corresponding author.}

\renewcommand{\shortauthors}{Chenxi Liu et al.}

\begin{abstract}
    Large Language Models (LLMs) have emerged as a promising paradigm for time series analytics, leveraging their massive parameters and the shared sequential nature of textual and time series data. However, a cross-modality gap exists between time series and textual data, as LLMs are pre-trained on textual corpora and are not inherently optimized for time series. In this tutorial, we provide an up-to-date overview of LLM-based cross-modal time series analytics. We introduce a taxonomy that classifies existing approaches into three groups based on cross-modal modeling strategies, e.g., conversion, alignment, and fusion, and then discuss their applications across a range of downstream tasks. In addition, we summarize several open challenges. This tutorial aims to expand the practical application of LLMs in solving real-world problems in cross-modal time series analytics while balancing effectiveness and efficiency. Participants will gain a thorough understanding of current advancements, methodologies, and future research directions in cross-modal time series analytics.
\end{abstract}




\maketitle

\section{Introduction}
With the proliferation of edge devices and the development of mobile sensing techniques, a large amount of time series data has been generated, enabling a variety of real-world applications~\cite{shang2023spatiotemporal,cai2024modeling}. Time series data typically takes the format of sequential observations with varying features~\cite{DBLP:journals/corr/abs-2401-17350,DBLP:conf/dsaa/AlnegheimishNBV24}. Considerable research efforts have been made to design time series analytics methods, which enable different downstream tasks, such as traffic forecasting~\cite{liu2024spatial}, disease classification~\cite{zhang2024dualtime}, and anomaly detection~\cite{xu2024pefad}.

Large language model (LLM)-based methods~\cite{touvron2023llama,radford2019language} have emerged as a new paradigm for time series analytics tasks. LLMs such as GPT~\cite{radford2019language}, LLaMA~\cite{touvron2023llama}, and DeepSeek~\cite{bi2024deepseek} exhibit strong modeling capacity due to their robust representation learned from vast and diverse language corpora, which enables them to encode complex patterns across various time series tasks~\cite{chenxi2021study,miao2024less,miao2025parameter}. In contrast to small classical models like recurrent neural network~\cite{DBLP:conf/hpcc/ChenWL20}, convolutional neural network~\cite{shen2018stepdeep}, or graph neural network-based methods~\cite{liu2024mvcar,lin2024hierarchical,xia2025prost} that are shallow due to limited by their relatively small number of learnable parameters, LLMs benefit from large-scale self-supervised pretraining and prompting strategies, allowing better transferability to downstream tasks~\cite{zhang2025llm,zhang2023controlling}.

Recent studies are inspired by time series and natural text exhibit similar formats (i.e., sequence), and assume that the generic knowledge learned by LLMs can be easily transferred to time series~\cite{bian2024multi,liu2025timekd}.
Although existing studies have introduced broad overviews of LLM-based time series methods~\cite{jin2024position,DBLP:conf/ijcai/0001C0S24,jiang2024empowering,DBLP:conf/kdd/LiangWNJ0SPW24}, they overlook the critical challenge posed by the cross-modality gap~\cite{liu2024timecma} between time series and textual data. To be specific, LLMs are pre-trained on textual corpora and are not inherently designed for time series, there is a pressing need to develop cross-modality modeling strategies that effectively integrate textual information into time series analytics.

\begin{figure}[t]
\centering
\includegraphics[width=0.45\textwidth]{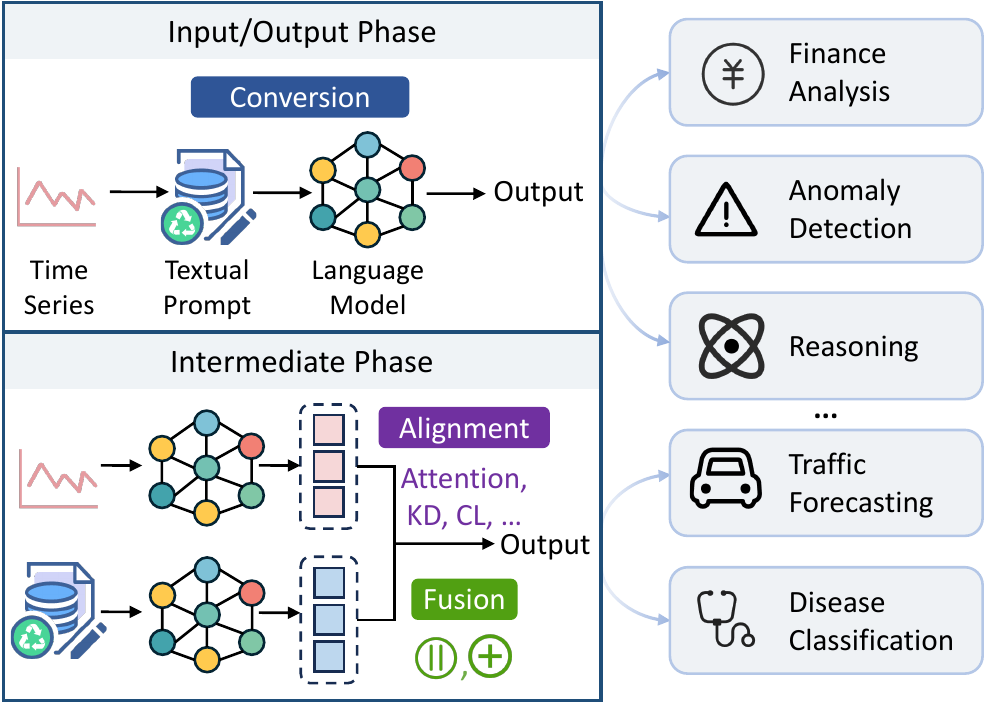}
\caption{Cross-Modal Time Series Analytics. Three cross-modal strategies: conversion, alignment, and fusion. Time series is converted to text in the input/output phase. In the intermediate phase, both modalities are embedded, followed by attention, knowledge distillation (KD), or contrastive learning (CL) for alignment; addition or concatenation for fusion.}
\label{fig:intro}
\end{figure}

This tutorial makes a unique contribution to the existing literature by addressing the cross-modality gap between time series and textual data, thereby enhancing LLM-based time series analytics. Figure~\ref{fig:intro} shows a general framework for cross-modal time series analytics via three strategies: (1) the \textbf{Conversion} strategy aims to switch time series to textual data at the input or output phase. (2) the \textbf{Alignment} strategy focuses on learn the association of time series and textual data via retrieval, contrastive learning, or knowledge distillation methods. (3) the \textbf{Fusion} strategy refers to the process of combining textual and time series data into a union representation. In addition, we discuss the applications and case studies across a range of time series tasks. Finally, we summarize the open challenges for future research. 

The remainder of this tutorial is organized as follows. Section 2 illustrates the relevance of this tutorial to SSTD. Section 3 describes the topics covered in the tutorial. Section 4 summarizes the open issues and future directions. Section 5 introduces an overlap statement of our previous related tutorials, surveys, and papers. Section 6 presents the tutors, short bio, and expertise. Section 7 contains a tutorial outline. Section 8 describes the intended audience.

\section{Relevance to the SSTD}
The International Symposium on Spatial and Temporal Databases (SSTD) has long been a leading forum for research in spatial and temporal data. With the rise of learning-based models, integrating LLMs into spatio-temporal analytics is timely and impactful. This tutorial bridges symbolic textual knowledge and structured time series data, addressing growing needs in mobility, urban computing, and environmental monitoring. By introducing LLM-enhanced cross-modal modeling strategies, it broadens SSTD’s methodological scope and fosters collaboration between database and AI researchers.


\section{Topics Covered}


\subsection{Modality Conversion}
Modality conversion transforms time series data into a textual modality at either the input or output phase, enabling direct interaction with LLMs. These studies typically involve wrapping time series values into text format, encoding temporal patterns as natural language descriptions, or abstracting sequences into instruction-style prompts~\cite{DBLP:conf/gis/0001VS22,wang2024news}. By converting time series to token sequences, LLMs can generate textual outputs for various applications, such as finance analysis~\cite{yu2023temporal}, anomaly detection~\cite{qiu2024ef}, and reasoning~\cite{zhang2023insight}. The modality conversion is particularly effective in zero-shot or few-shot settings, where domain-specific priors can be embedded in prompt templates to guide generation. LLMTime~\cite{DBLP:conf/nips/GruverFQW23}, where time series is serialized as numerical strings for prediction, which augments time series forecasting with natural language instructions and explanations. PromptCast~\cite{xue2023promptcast} transforms the numerical input and output into prompts, and the time series forecasting is framed in a sentence-to-sentence manner.

\subsection{Cross-Modal Alignment}
Cross-modal alignment seeks to bridge the semantic gap between time series and textual modalities by aligning their latent representations. This typically occurs in the intermediate phase, where both modalities are separately encoded and then matched via attention-based retrieval, contrastive learning, or knowledge distillation. 

Retrieval uses one modality to access relevant information in another and can be categorized as unidirectional and bidirectional. In unidirectional retrieval, TimeCMA~\cite{liu2024timecma} designs a similarity-based retrieval to align the outputs of the time series branch and LLM branch with the time series representation, using the time series embeddings as queries.
Bidirectional retrieval extends unidirectional retrieval by allowing both time series and textual embeddings to retrieve information from each other. 
For instance, LeRet~\cite{huang2024leret} introduces a bidirectional retrieval for time series forecasting. Instead of relying solely on time series embeddings as queries, LeRet allows textual embeddings to retrieve relevant time series features, creating a dynamic exchange between the two modalities.

Contrastive learning establishes a shared representation by maximizing the agreement between time series and textual embeddings while minimizing the similarity between non-corresponding pairs~\cite{DBLP:conf/iclr/OzyurtF023}. For example, Chen et al.~\cite{chen2024llm} propose a contrastive module to align time series and textual prompts by maximizing the mutual information. 
HiTime~\cite{tao2024hierarchical} presents a dual-view contrastive alignment module that bridges the gap between modalities via facilitating semantic space alignment between time series and contextual prompt. 
TEST~\cite{DBLP:conf/iclr/Sun0LH24} builds an encoder to embed TS via instance-wise, feature-wise, and text-prototype-aligned contrast, where the TS embedding space is aligned to LLM’s embedding layer space.

Knowledge distillation (KD) achieves a small student model from an LLM, enabling efficient inference solely on the distilled student model~\cite{li2024self}.
Recent works have been proposed to address the misalignment problem via KD, which can generally be categorized into black-box distillation~\cite{liu2024large} and white-box distillation\cite{DBLP:conf/iclr/Gu0WH24} based on the accessibility of the teacher model’s internal information during the distillation process.
TimeKD~\cite{liu2025timekd} employs privileged knowledge distillation to align the time series and textual modality, which includes correlation and feature distillations, enabling the student model to learn the teacher’s behavior while minimizing the output discrepancy between them.
CALF~\cite{liu2024taming} is the representative black-box KD method that aligns LLMs for time series forecasting via cross-modal fine-tuning. 


\subsection{Cross-Modal Fusion}
Cross-modal fusion integrates time series and textual modalities into a unified representation. Fusion can occur at early, middle, or late stages, depending on the level of abstraction at which modalities are combined. In early fusion, raw features are concatenated before encoding; in middle fusion, hidden embeddings are merged; and in late fusion, prediction logits are aggregated. 

Addition-based fusion integrates textual embeddings with time series by summing their feature vectors. This method allows models to incorporate textual information without increasing the dimensionality of the feature space, making it a computationally efficient alternative to concatenation. Unlike concatenation, addition-based fusion maintains a compact representation, ensuring that the model does not introduce unnecessary complexity while still leveraging multimodal information.
Several studies have adopted addition-based fusion for time series analysis. Time-MMD~\cite{liu2024timemmd}, GPT4MTS~\cite{jia2024gpt4mts}, AutoTimes~\cite{liu2024autotimes}, and T3~\cite{han2024event} add the textual embedding with time series embedding for analysis, respectively.

Concatenation-based fusion directly merges textual embeddings with time series features to create a joint representation. 
While concatenation provides a straightforward way to multimodal integration, it can increase the dimensionality of the feature space, leading to greater computational complexity. Moreover, the lack of explicit alignment mechanisms between modalities may introduce noise, reducing the effectiveness of downstream tasks.
For instance, UniTime~\cite{liu2024unitime} concatenates the contextual prompt embedding with time series embedding to retain a LLM-based unified model for cross-domain time series forecasting.

\section{Open Issues and Future Directions}
\subsection{Open Issues}

\textbf{Improving Effectiveness.} While LLM-based cross-modality methods have demonstrated strong capabilities, they do not always surpass smaller, task-specific models~\cite{DBLP:conf/nips/WangWDQZLQWL24}. Employing LLMs with an excessive number of parameters can lead to overfitting or hallucination, particularly on specialized tasks. Future research could focus on techniques such as multi-agent LLMs collaboration that could help improve model effectiveness by allowing models to adjust to changing data distribution.

\noindent \textbf{Efficient Optimization.} Despite their success, existing studies still meet the challenge of high computational costs, particularly when processing long sequences, more tokens, or handling multivariate data. This is due to the high dimensionality of multivariate time series (i.e., multiple variables over multiple timestamps) and the multi-head attention mechanism within LLMs
Recent advancements have explored strategies to mitigate this challenges, such as time series patching~\cite{bian2024multi}, last token storage~\cite{liu2024timecma}, knowledge distillation~\cite{DBLP:conf/iclr/Gu0WH24}. Future research could focus on developing lightweight architectures, efficient attention mechanisms, and adaptive computation frameworks to optimize efficiency and scalability.

\noindent \textbf{Transparency of LLMs.} LLMs have demonstrated remarkable performance in textual–time series analytics~\cite{wang2024news}, yet they often operate as “black-box” systems, raising concerns about their reasoning processes and transparency. Much of the current research primarily applies or fine-tunes LLMs without an explicit focus on exposing their internal reasoning processes. This lack of interpretability can hinder trust, particularly in high-stakes applications such as healthcare and finance. Moreover, LLMs are prone to generating hallucinations, seemingly plausible but incorrect outputs. Future research on textual–time series analysis could prioritize enhancing the transparency of LLMs, ensuring that these models operate more reliably during subsequent alignment or fusion processes.

\subsection{Future Directions}

\textbf{Multimodal LLMs.} The Multimodal LLMs extend traditional text-based large language models by integrating diverse modalities, such as numerical time series, images, and sensor data, into a unified representation space. Leveraging multimodal LLMs in cross-modal time series analytics allows for richer contextual understanding and more robust predictive performance. Research efforts could focus on developing multimodal alignment techniques to integrate heterogeneous data and designing multimodal pre-training strategies to capture spatial and temporal dynamics.

\noindent \textbf{Multi-Agent LLMs.} The Multi-Agent LLMs involve multiple specialized language models collaboratively performing analytical tasks, each tailored to different data modalities or subtasks in cross-modal analytics. In cross-modal time series analytics, deploying multiple LLM agents allows each to specialize in distinct areas. These agents can dynamically interact, exchanging intermediate insights and refining their analyses through iterative communication and collaborative reasoning. Future work may emphasize developing efficient communication protocols and coordination methods for conflict resolution among agents.

\noindent \textbf{RAG-Enhanced Large Models.} Integrating Retrieval-Augmented Generation (RAG) into large models enhances their ability to handle cross-modal time series analytics by enabling real-time access to external knowledge bases and historical data. RAG-enhanced models dynamically retrieve contextual information from time series databases or textual documents. Future research could investigate indexing methods for efficient information extraction and adaptive retrieval strategies capable of learning from user interactions.

\section{Overlap Statement}

\textbf{Related Tutorials from Our Organizers:}
\begin{itemize}
    \item \textbf{Web-Centric Human Mobility Analytics: Methods, Applications, and Future Directions in the LLM Era (WWW\\'25 Tutorial~\cite{www25tutorial})}: This tutorial held at WWW'25 in Sydney, Australia, April 29, 2025. The tutorial offers an in-depth look at web-centric human mobility analytics, organized according to three levels: location-level, individual-level, and macro-level. The tutorial encompasses cutting-edge learning frameworks such as federated learning as well as continual learning and innovative applications of LLMs, which enhances predictive analytics and expands the capabilities of mobility analysis.
    \item \textbf{Multimodal Learning for Spatio-Temporal Data Mining (MM'25 Tutorial~\cite{mm25tutorial})}: This tutorial will hold at MM'25 in Dublin, Ireland, November, 2025. This half-day tutorial will explore how multimodal learning can transform spatio-temporal analysis. Topics will include fundamentals of spatio-temporal mining, challenges of integrating heterogeneous data state-of-the-art multimodal modeling techniques, and emerging research trends.
\end{itemize}

\noindent \textbf{Related Surveys by the Presenters}
\begin{itemize}
    \item Survey on cross-modality modeling for time series analytics: IJCAI'25~\cite{liu2025ijcai}.
    \item Survey on spatio-temporal trajectory computing: CSUR'25~\cite{liu2025csur}.
    \item Survey on neural open information extraction: IJCAI'22~\cite{zhou2022survey}.
\end{itemize}

\noindent \textbf{Related Papers by the Presenters}
\begin{itemize}\setlength{\itemsep}{2pt}\setlength{\parsep}{0pt}
    \item Large Language Models: 
    TKDE'25~\cite{liu2025stllm_plus}, ICDE'25~\cite{liu2025timekd}, AAAI'25\\\cite{liu2024timecma}, 
    TMC'25~\cite{miao2025parameter},
    SIGKDD'25~\cite{xu2024pefad},
    MDM'24~\cite{liu2024spatial}.
    \item Time Series Analytics:
    PVLDB'25~\cite{miao2024less}, 
    SIGKDD'25~\cite{zhao2024unsupervised},
    TKDE'25\\\cite{miao2025spatio}, ICDE'24~\cite{miao2024unified}, IJCAI'24~\cite{zhang2024score}, ICLR'24~\cite{hettige2024airphynet}, 
    TKDE'24~\cite{lai2024lightcts},
    SIGMOD'23~\cite{lai2023lightcts},
    AAAI'23~\cite{zhang2023autostl}, CIKM'22~\cite{wang2022generative}, TKDE'22~\cite{miao2022mba}, TKDE'22~\cite{li2022fine}, CIKM'22~\cite{xu2022traffic}, SDM'21~\cite{wang2021mt}, AAAI'20~\cite{li2020tensor}.
    \item Spatio-Temporal Mobility Modeling: ICDE'25~\cite{miao2025traj}, AAAI'25\\\cite{wang2025c2f}, SIGKDD'24~\cite{zhu2024controltraj}, 
    SIGKDD'24~\cite{wang2024multi},
    ICDE'24~\cite{wang2024collectively}, ICDE'24~\cite{liu2024lighttr}, ICDE'23~\cite{zhang2023online}, ECML-PKDD'22~\cite{zhang2022multi}, ICDE'22~\cite{ruan2022discovering}, SIGKDD'21~\cite{wang2021error}.
    \item Cross-modal learning: AAAI'25~\cite{liu2024timecma}, ICDE'25~\cite{liu2025timekd}, TMM'23\cite{liu2023relation}.
\end{itemize}

\section{Tutors, Short Bio, and Expertise}

\textbf{Presenter \#1: Panos Kalnis} is a Professor of Computer Science at the King Abdullah University of Science and Technology (KAUST), and served as Chair of the Computer Science program from 2014 to 2018. He was involved in the designing and testing of VLSI chips and worked in several companies on database designing, e-commerce projects, and web applications. 
His research interests include Big Data, Parallel and Distributed Systems, Large Graphs, and Systems for Machine Learning. 

\vspace{1mm}

\noindent \textbf{Presenter \#2: Cheng Long} is an Associate Professor at the College of Computing and Data Science (CCDS), Nanyang Technological University (NTU), Singapore. He has received several awards, most recently a nomination for the Best Paper Award at SIGMOD 2020. I serve as an Associate Editor for the Data \& Knowledge Engineering Journal and have held roles such as Industry and Application Track Co-Chair for MDM 2024 and PhD Symposium Co-Chair for ICDE 2022. He has research interests in data management and data mining. Specifically, he works in high-dimensional vector data management, spatial data management with machine learning-based techniques, spatial data mining in the urban domain, and graph data mining. 

\vspace{1mm}

\noindent \textbf{Presenter \#3: Yan Zhao} a Professor with University of Electronic Science and Technology of China. Her research interests include spatio-temporal databases, trajectory computing, spatial crowdsourcing, recommendation systems, as well as many other data-driven applications. She has received a few best paper awards at international conferences including DASFAA 2023, APWeB-WAIM 2022, and MDM 2022. She has received the ACM SIGSPATIAL China Chapter Doctoral Dissertation Award in 2021 and the Excellent Doctoral Thesis of Jiangsu Province, China, in 2022.

\vspace{1mm}

\noindent \textbf{Presenter \#4: Ziyue Li} is an Assistant Professor with Technical University of Munich. He is also the chief machine learning scientist with EWI. His research targets high-dimensional data mining and deep learning methodologies for real-world spatio-temporal problems. His expertises are tensor analysis, spatio-temporal data, and statistical machine learning. Those methods have been applied to various industries, mainly in smart transportation, as well as in smart manufacturing and multimedia. His works have been awarded various Best Paper awards in INFORMS, IISE, and CASE.

\vspace{1mm}

\noindent \textbf{Presenter \#5: Chenxi Liu} is a Research Fellow at Nanyang Technological University (NTU). Her research interests include time series analytics, large language models, and multimodal learning. She has been recognized with several international awards, including the SoBigData Award for Diversity and Inclusion funded by the European Union, the Visit Grant from the Danish Data Science Academy, the IJCAI-25 Travel Grant, a CNCC Travel Grant from the China Computer Federation, and a WiEST Development Grant from the Women in Engineering, Science, and Technology, NTU.


\vspace{1mm}

\noindent \textbf{Presenter \#6: Hao Miao} is a Research Assistant Professor at The Hong Kong Polytechnic University with research interests in spatio-temporal data analytics, trajectory computing, and spatial crowd sourcing. He has published over 20 research papers in top-tier venues (e.g., PVLDB, SIGKDD, ICDE and TKDE). He has received a PVLDB 2025 Travel Award, SIGIR CIKM 2020 Travel Award and an Otto Moensted Foundation Study Abroad Award.

\section{Tutorial Outline}
The tutorial is designed to be a 90-minute session:

\noindent \textbf{Section 1. Welcome and Introduction (15 mins)}

\noindent \textbf{Presenter: Panos Kalnis}

\noindent Introduces the background of time series data, foundation models, and pre-trained LLMs. 

\vspace{1mm}

\noindent \textbf{Section 2. LLMs for Time Series (20 mins)}

\noindent \textbf{Presenter: Hao Miao}

\noindent Covers the evolution of LLMs and case studies in applying LLMs to non-textual modalities like time series data.

\vspace{1mm}

\noindent \textbf{Section 3. Cross-Modal Modeling (25 mins)}

\noindent \textbf{Presenter: Chenxi Liu}

\noindent Introduces a taxonomy covering three strategies: (i) modality conversion, (ii) cross-modal alignment, especially TimeCMA~\cite{liu2024timecma} and TimeKD~\cite{liu2025timekd}, and (iii) cross-modal fusion.

\vspace{1mm}





\noindent \textbf{Section 5. Open Challenges and Opportunities (10 mins)}

\noindent \textbf{Presenter: Yan Zhao}

\noindent Discusses key research challenges, including model effectiveness, efficiency, and transparency.

\noindent \textbf{Section 6. Future Directions (10 mins)}

\noindent \textbf{Presenter: Ziyue Li}

\noindent Propose future directions, including multimodal LLMs, multi-agent LLMs, and RAG-enhanced large models.

\vspace{1mm}

\noindent \textbf{Section 7. Q\&A and Wrap-up (10 mins)}

\noindent \textbf{Presenter: Cheng Long}

\noindent Recaps the key insights from the tutorial and engages the audience with Q\&A.

\section{Intended Audience}

This tutorial is designed for a broad audience of researchers who are interested in time series analytics, LLMs, and multimodal learning. We assume participants have a basic understanding of machine learning and sequential data modeling, making the content accessible to professionals and students alike. Attendees will gain a comprehensive understanding of the methodologies and challenges in cross-modal time series analytics. They will also become familiar with emerging applications across domains, and will learn how to integrate textual knowledge into time series tasks using LLMs.

\bibliographystyle{ACM-Reference-Format}
\bibliography{sample-base}

\end{document}